\documentclass{article}

\usepackage{arxiv}

\usepackage[utf8]{inputenc} 
\usepackage[T1]{fontenc}    
\usepackage[hidelinks]{hyperref}       
\usepackage{url}            
\usepackage{booktabs}       
\usepackage{amsfonts}       
\usepackage{nicefrac}       
\usepackage{microtype}      
\usepackage{lipsum}		
\usepackage{graphicx}
\usepackage{doi}
\usepackage{amsmath}

\title{Imputation of Missing Data in Smooth Pursuit Eye Movements Using a Self-Attention-based Deep Learning Approach}

\author{ \href{https://orcid.org/0000-0002-7668-3413}{\includegraphics[scale=0.06]{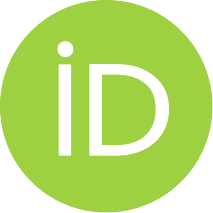}\hspace{1mm}Mehdi Bejani}\\
	ETSI Telecomunicación\\ Universidad Politécnica de Madrid\\ 28040, Madrid, Spain\\
	\texttt{mehdi.bejani@upm.es} \\
	\And
	Guillermo Perez-de-Arenaza-Pozo \\
	ETSI Telecomunicación\\ Universidad Politécnica de Madrid\\ 28040, Madrid, Spain\\
	\texttt{g.perezdearenaza@alumnos.upm.es} \\
	 \And
	 \href{https://orcid.org/0000-0002-1928-773X}{\includegraphics[scale=0.06]{orcid.pdf}\hspace{1mm}Julián D. Arias-Londoño} \\
	ETSI Telecomunicación\\ Universidad Politécnica de Madrid\\ 28040, Madrid, Spain\\
	 \texttt{julian.arias@upm.es} \\
	 \And
	 \href{https://orcid.org/0000-0001-7348-3291}{\includegraphics[scale=0.06]{orcid.pdf}\hspace{1mm}Juan I. Godino-LLorente}\thanks{Corresponding author. All authors contributed in the same way.}  \\
	ETSI Telecomunicación\\ Universidad Politécnica de Madrid\\ 28040, Madrid, Spain\\
	 \texttt{ignacio.godino@upm.es}
}

\renewcommand{\shorttitle}{Imputation of Missing Data in Smooth Pursuit Eye Movements}

\hypersetup{
pdftitle={Imputation of Missing Data in Smooth Pursuit Eye Movements},
pdfsubject={Smooth Pursuit Eye Movements},
pdfauthor={M.~Bejani, G.~Perez-de-Arenaza-Pozo, J.D. ~Arias-Londoño, J.I. ~Godino-Llorente},
pdfkeywords={Imputation, Smooth Pursuit Eye Movements, SAITS},
}

\begin{document}
\maketitle

\begin{abstract}
Missing data is a relevant issue in time series, especially in biomedical sequences such as those corresponding to smooth pursuit eye movements, which often contain gaps due to eye blinks and track losses, complicating the analysis and extraction of meaningful biomarkers. In this paper, a novel imputation framework is proposed using Self-Attention-based Imputation networks for time series, which leverages the power of deep learning and self-attention mechanisms to impute missing data. We further refine the imputed data using a custom made autoencoder, tailored to represent smooth pursuit eye movement sequences. The proposed approach was implemented using 5,504 sequences from 172 Parkinsonian patients and healthy controls. Results show a significant improvement in the accuracy of reconstructed eye movement sequences with respect to other state of the art techniques, substantially reducing the values for common time domain error metrics such as the mean absolute error, mean relative error, and root mean square error, while also preserving the signal’s frequency domain characteristics. Moreover, it demonstrates robustness when large intervals of data are missing. This method offers an alternative solution for robustly handling missing data in time series, enhancing the reliability of smooth pursuit analysis for the screening and monitoring of neurodegenerative disorders.
\end{abstract}

\keywords{Imputation \and Smooth Pursuit Eye Movements \and SAITS \and Blinking \and Missing data \and Autoencoder\and Self-Attention-based Imputation for Time Series.}

\section{Introduction}
\label{sec:introduction}
Missing values are particularly prevalent in time series collected from physiological phenomena, and usually arise from various causes, including sensor malfunctions, transmission errors, or signal disruptions during data acquisition. These missing data pose significant challenges for the accurate analysis and interpretation of the underlying sequences, significantly limiting the interpretability of the sequences. Therefore, imputing missing values is a critical and essential step before any modelling or analysis is conducted. However, traditional methods of handling missing data, such as linear and polynomial interpolation, mean imputation, and complete-case analysis, fail to capture the complex temporal dependencies in time series \cite{b1}\cite{b2}\cite{b3}\cite{b4}.  

On the other hand, the analysis of Smooth Pursuit Eye Movements (SPEM) holds significant diagnostic potential for many neurodegenerative diseases, yet it is frequently hampered by the presence of missing values \cite{b5}\cite{b6}\cite{b7}\cite{b8}\cite{b9}\cite{b10}. Such gaps in the data can arise from various sources, including blinks, occlusions, technical malfunctions, tracking losses (due to movements of the patient, changes in the illumination conditions, the use of intraocular lenses, etc), and other unforeseen situations. Developing robust methods to handle incomplete SPEM sequences is paramount to fully realize the potential of eye movement analysis in clinical diagnostics. 

Literature reports a large number of works dealing with the complex task of imputing missing values from physiological signals. To this respect, deep learning based approaches have emerged in recent years as effective solutions for handling missing values in physiological data. Generative models, such as diffusion approach models and Generative Adversarial Networks (GAN), have been widely explored for signal imputation, denoising, and generation. For instance, \cite{b11} proposed a diffusion-based time series imputation framework using structured state space models, addressing missing data in electrocardiographic (ECG) signals. This diffusion approach models the uncertainty of the missing points, providing a probabilistic solution for both imputation and forecasting. In a related line of work, DiffECG \cite{b12} employs a probabilistic diffusion model for ECG signal synthesis, offering a versatile method for not only imputing missing values but also generating realistic ECG signals. Additionally, the seminal work in \cite{b13} introduced GAIN, a generative adversarial network-based approach for missing data imputation. GAIN has been applied successfully to various domains, including physiological signals, demonstrating its effectiveness in capturing complex data distributions and generating high-quality imputations. In addition, \cite{b14} introduced the use of GAN for the imputation of ECG signals, leveraging the CycleGAN framework for translation and imputation tasks. CycleGAN has demonstrated its utility in tasks requiring mapping between domains without paired data, making it particularly useful in contexts where complete ECG recordings are unavailable. In another study, GAN-based methods were applied to electroencephalographic (EEG) signals, focusing on contextual imputation in scenarios with missing sequences \cite{b15}. These methods underscore the adaptability of GAN in handling different types of physiological signals, including EEG and ECG, by learning to generate realistic signal patterns from incomplete datasets. Other recent approaches explore the periodic nature of physiological signals. The work by \cite{b16} proposed a Fourier-based decoder within a conditional generative model, specifically designed for generating periodic signals. This approach highlights the importance of leveraging signal periodicity in improving the quality of the generated or imputed data, particularly in long, continuous time series like ECG and EEG. A broader review of deep generative models for physiological signals was conducted by \cite{b17}, where the authors systematically evaluated various architectures, including Long Short-Term Memory networks (LSTM), GAN, and variational autoencoders (VAE). The study highlights how different deep learning architectures perform in tasks such as signal synthesis, denoising, and imputation, with LSTM being commonly employed for handling the sequential nature of physiological data. In summary, the ongoing advancements in deep learning, particularly through the use of generative models, have provided powerful tools for addressing missing data in physiological signals. These models not only impute missing data but also contribute to the synthesis and denoising of signals, thereby improving the quality of the data available for analysis in medical applications. 

\begin{figure}[!t]
  \centering
  \includegraphics[scale=0.8]{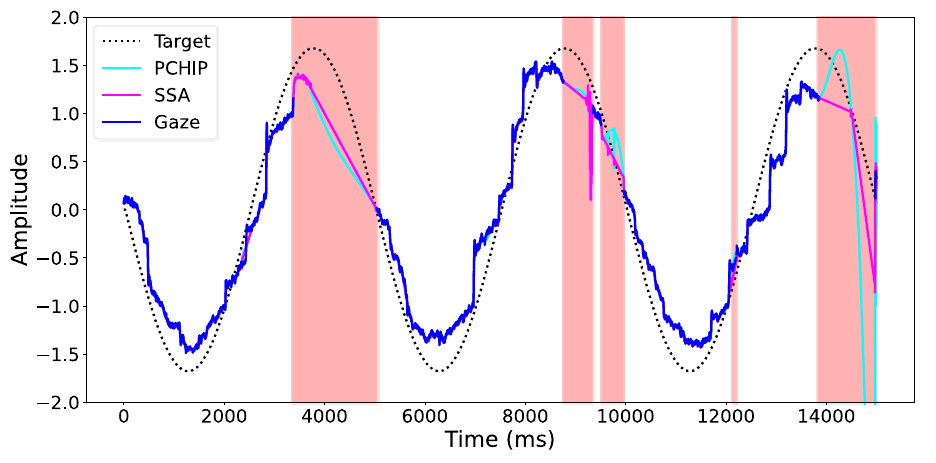} 
  \caption{Illustration of an SPEM sequence (blue) along with the target (dotted black) for participant HG032\_V2, SPT12, y-axis and right eye. Red stripes correspond to missing segments. Imputed data using the PCHIP method is represented in light blue, while data imputed using the SSA method is shown in purple.}
  \label{fig:fig1}
\end{figure}

In the specific case of missing data imputation of SPEM sequences, \cite{b18} used a Piecewise Cubic Hermite Interpolating Polynomial (PCHIP) \cite{b19} method. In a subsequent study, \cite{b20} used a recurrent forecasting method based on the Singular Spectrum Analysis (SSA) technique \cite{b20}, \cite{b21}. Although SSA showed improvements over PCHIP, its performance declined significantly in cases with a large number of missing data. Fig. \ref{fig:fig1} highlights the limitations of both PCHIP and SSA methods imputing missing values (highlighted as red stripes), especially in those cases with long missing segments. 

Statistical and traditional imputation methods, such as SSA and PCHIP, often struggle to handle the complexities of time series data in those scenarios involving long missing intervals and nonlinear patterns. These methods primarily rely on local information and make linear or stationary assumptions about the underlying data, which limits their ability to capture long-term dependencies and intricate temporal dynamics. For example, SSA decomposes the time series into components based on principal components analysis, assuming stationarity, which fails to account for nonstationary behaviours common in certain physiological signals, like SPEM. Similarly, PCHIP interpolates based on neighbouring data points, which can lead to significant inaccuracies when missing data spans large intervals. These limitations make such methods less effective for imputing time series data with high variability and complex interactions. Thus, more advanced imputation methods are required to model temporal dependencies and provide accurate reconstructions of missing segments in SPEM sequences \cite{b22}, \cite{b23}, \cite{b24}, \cite{b25}. 

In this paper, we use a procedure based on the Self-Attention-based Imputation for Time Series (SAITS) \cite{b26} method, which is based on a cutting-edge deep learning framework. Deep learning-based imputation methods, particularly the SAITS model, offer a powerful alternative to other methods by leveraging self-attention mechanisms to dynamically model both global temporal structures and local dependencies. Unlike traditional approaches, SAITS does not impose a strict left-to-right sequential assumption (like ARIMA or LSTM models), allowing it to efficiently capture complex inter-variable relationships and long-term patterns in time series data. 

SAITS has demonstrated state-of-the-art performance in time series imputation tasks across various benchmarks, outperforming traditional models and other deep learning methods \cite{b11}. In contrast to generative models like GAN, which often encounter challenges such as mode collapse and convergence difficulties, SAITS employs the Transformer architecture to address time series imputation. Traditional recurrent neural networks can struggle with capturing long-term dependencies due to issues like vanishing gradients and sequential bottlenecks. However, SAITS has the potential to overcome these limitations by leveraging self-attention mechanisms, allowing the model to capture both local and global temporal dependencies and feature correlations across all time steps simultaneously. 

By incorporating SAITS into this study, we aim to enhance the robustness and accuracy of the imputation of missing data in SPEM recordings. And, building on this model, we have additionally introduced a new step using a refinement autoencoder (RAE) to further improve the quality of the imputed signals. The combination of SAITS with the customised RAE is used to reconstruct the missing sequences in a more precise way, ensuring that both global temporal structures and local signal nuances are effectively captured and tailored to the unique characteristics of SPEM sequences.

The remainder of this paper is structured as follows: Section II presents the materials used in this study, including details on the participants, clinical assessments, and data collection protocols. Section III describes the methodology, explaining the steps in our proposed imputation pipeline, including downsampling, SAITS-based imputation, upsampling, and refinement using the RAE. Section IV discusses the experimental results, highlighting the effectiveness of the imputation pipeline through various evaluation metrics. Finally, Section V provides conclusions, discusses the implications of our findings, and outlines potential future work.

\section{MATERIALS}
 This section presents information about participants and clinical evaluations, as well as the recording methods and explorations used in the study. Explorations are based on the Smooth Pursuit paradigm.

\subsection{Participants and clinical assessments}
 Forty-one patients with Parkinson’s Disease (PD) and forty-seven healthy control subjects (Ctrl) of the same age and sex were initially recruited. The average age of the patients with PD was 61.80 years (range 47-78 years) and 63.62 years (range: 45–79 years) for the Ctrl group. Additionally, twelve young healthy control subjects were included, with an average age of 25.04 years. 

Participants were recorded in two different sessions separated 24 months, but following the same protocol. However, due to different reasons, some participants from the first round of recordings were unable to attend the second session. These reasons included personal decisions to withdraw from the study, physical conditions that prevented participation (such as recent cataract surgery, leg ulcers, or hospitalisations), or the impossibility of re-establishing contact with the participants.

The second session includes 72 participants: 39 with PD, and 33 Ctrl subjects. The average age of participants with PD was 64.63 years (range 48-79), while for the Ctrl it was 64.48 years (range 46-80), slightly differing due to the exclusion of some participants in the second session. Since the primary objective of this study is to impute missing data, all participants were processed collectively, assuming independence between observations. Therefore, the final dataset includes 172 participants, 80 with PD and 92 Ctrl. subjects.

Participants received a clinical evaluation, and a comprehensive medical history was recorded, including the age of the onset of PD, the duration of the disease, symptoms, and complications. The Unified Parkinson's Disease Rating Scale (UPDRS), the Montreal Cognitive Assessment (MoCA), and the Beck Depression Inventory were used to evaluate patients with PD. The mean UPDRS for the PD group was 16.88, while it was 1.5 for the Ctrl group. 

The corpus was collected over three years in two hospitals in Madrid, Spain: Fuenlabrada Hospital, and Gregorio Marañón Hospital. The Ethics Committee of the Fuenlabrada and Gregorio Marañón Hospitals approved this research. All participants received written informed consent and signed participation forms.

\subsection{Recording and tasks}

 Eye movements were recorded with an EyeLink 1000 Plus infrared video-based binocular (SR Research Ltd, Ontario, Canada) sampling at 1 kHz. Two computers drive the process: one controls the eye tracking system, while the other presents the stimuli. An LED monitor was placed 60 cm in front of the patient, which was attached to the second computer.

\begin{figure*}[!t]
  \centering
a{\includegraphics[width=0.23\textwidth]{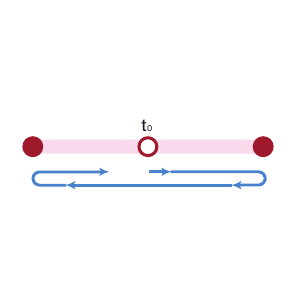}\label{fig:stim_a}}
  \hfill
 b{\includegraphics[width=0.23\textwidth]{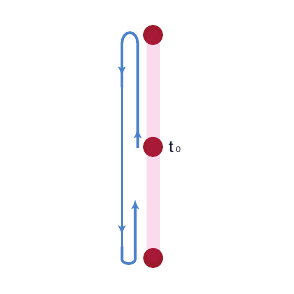}\label{fig:stim_b}}
  \hfill
 c{\includegraphics[width=0.23\textwidth]{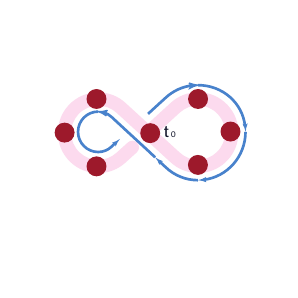}\label{fig:stim_c}}
  \hfill
d{\includegraphics[width=0.23\textwidth]{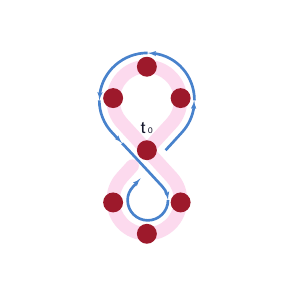}\label{fig:stim_d}}
  
  \caption{Visual stimuli used in the twelve SPT conditions. Participants initially fix on a central white dot for 200~ms before tracking a red dot moving in specific patterns.
  (a) horizontal linear motion (SPT1--SPT4); (b) vertical linear motion (SPT5--SPT8); (c) horizontal lemniscate (SPT9--SPT10); (d) vertical lemniscate (SPT11--SPT12). Blue arrows indicate the motion direction \cite{b20}.}
  \label{fig:fig2}
\end{figure*}

Participants were seated with their heads resting on a chin rest. The distance from the upper knob of the camera to the front of the chin rest is 50 cm. Before each recording session, the eye tracker system was calibrated using a 9-point grid that covers the area where the targets were presented. The stimuli start with a white dot in the center of the screen. 200 ms later, a moving red-point appears on the screen, which represents the target to be followed by the patient. Twelve Smooth Pursuits Task (SPT) were performed horizontally, vertically and combining both axes (2-D), following the patterns described in Fig. \ref{fig:fig2}. Furthermore, for each direction, four tasks with various velocities and amplitudes were defined as in \cite{b20}. As expected, eye blinks lead to segments with missing data in a large number of recordings. We analyzed sequences along the x-axis for horizontal SPTs, the y-axis for vertical SPTs, and both x- and y-axes for 2D SPTs. With data collected from 172 participants, considering both left and right sequences for each individual, a total of 5,504 sequences were obtained. Among these, 3,112 sequences were complete, containing no missing values, while the remaining 2,392 sequences included missing values. 

\section{METHODS}
\begin{figure}[!t]
  \centering
  \includegraphics[scale=0.8]{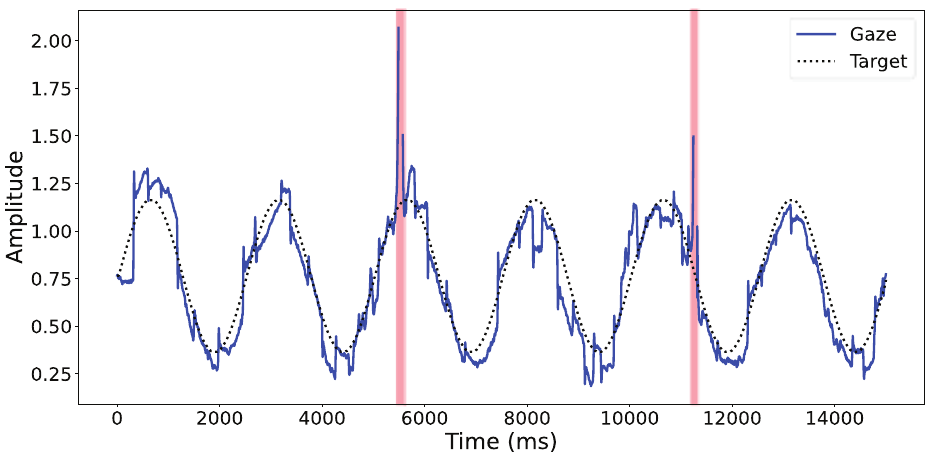} 
  \caption{Illustration of an SPEM sequence (in blue) along with the target (in dotted black) for participant HF021, SPT11, x-axis and left eye. Red stripes correspond with the missing segments.}
  \label{fig:fig3}
\end{figure}

For each SPT, the process starts with the identification and marking of the blinks to identify the missing data within the sequences. During blinks, the complete closure of the eyelids results in a temporary loss of the eye's position, leading to a pronounced artifact in the recorded sequence. As illustrated in Fig. \ref{fig:fig3}, this artifact manifests as an abrupt transition to infinity in the amplitude —recorded as NaN—, which rapidly returns to its baseline once the eyelids open again. 

To accurately identify and manage these artifacts, the procedure begins by scanning the recorded sequences along both the x- and y-axes to detect blocks of missing values. These blocks are annotated to mark the onset and offset of the blink. The blink span is further refined by extending the analysis to include the sharp signal changes immediately adjacent to these missing blocks, continuing until the slope of the signal stabilizes. In cases where consecutive blinks occur in close proximity, they are merged into a single, extended blink event to ensure continuity. This procedure is presented in \cite{b20}. A MATLAB\textsuperscript{®} custom software was developed for this purpose. 

\begin{figure*}[!t]
  \centering
  \includegraphics[scale=0.7]{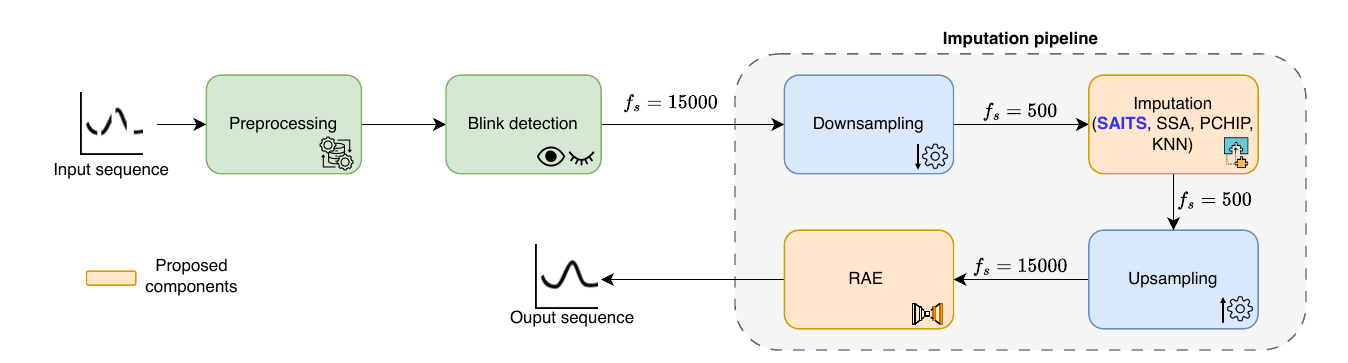}
  \caption{Illustration of the four-step pipeline followed for imputing missing data in SPEM Recordings.}
  \label{fig:fig4}
\end{figure*}

After blink detection and preprocessing, the imputation of the missing data due to blinks was carried out following a four-step pipeline: a) downsampling to reduce the number of points; b) imputation using the SAITS model; c) upsampling using a cubic interpolation to bring back the sequence to its original length; and, d) a refinement process to improve the accuracy of the reconstruction and to enhance the interpolated signals using a RAE. Fig. \ref{fig:fig4} illustrates this pipeline. Each of these steps is described in detail in the following sections.

\subsection{Downsampling}
Given the length of the recorded sequences (15,000 samples), a downsampling by a factor of 30 (i.e., up to 500 samples) was performed to make the process computationally feasible. This process introduces a low-pass filter in the sequence, removing the high-frequency components of the SPEM sequences. Thus, additional methods will later be required to restore the original characteristics of the signal. 

\subsection{Imputation}
The SAITS model provides certain advantages over traditional Transformer models and diffusion-based approaches like the Structured State Space Model for Diffusion (SSSD) \cite{b11}. By integrating two diagonally masked self-attention (DMSA) blocks, SAITS jointly captures temporal dependencies and inter-feature correlations, which leads to both higher imputation accuracy and faster convergence. Unlike vanilla Transformers —which may struggle with irregular sampling and variable-length gaps— SAITS leverages its masking scheme to prevent each time step from attending to itself, thereby ensuring that missing values are inferred from the full global context rather than only adjacent observations. Diffusion methods like SSSD, while powerful, incur significantly greater computational overhead and require complex multistage training, making SAITS a more efficient choice for real-time or resource-constrained applications.

At its core, SAITS comprises the DMSA blocks and a subsequent Weighted Combination block (WCB), which fuses multiple imputation hypotheses into a single refined output. The DMSA layers enforce diagonal masking so that each position must draw information exclusively from other time points, enhancing the model’s ability to learn long‑range patterns and to mitigate error propagation common in autoregressive approaches. Furthermore, the self‑attention mechanism dynamically assigns importance weights according to both the learned feature representations and the current missingness mask, improving flexibility when handling extended gaps in the data. This capability is particularly advantageous for the SPEM recordings, where complex, nonstationary dynamics and long‑term correlations are present. On the other hand, the WCB plays a crucial role in aggregating the outputs from the DMSA blocks to produce the final imputation. It dynamically assigns weights to the learned representations, emphasizing the most informative features and thereby enhancing the accuracy of the imputation.  

One single model was created for all the sequences recorded in the dataset (i.e., for all SPTs). This approach is carried out for the sake of simplicity and generalizability, as the model is optimised to effectively reconstruct missing data across various scenarios without the need for group-specific tailoring. 

The SAITS network was trained on z‑score‑normalised SPEM sequences that had been downsampled to 500 samples and corrupted with artificial missing values (see Section \ref{sec:gen_blinks}). Recordings from 137 participants formed the training pool, while data from 35 participants were held out for final evaluation. The training set was subjected to 10‑fold cross‑validation: for each fold, the model —comprising two DMSA blocks, four attention heads, a 256‑dimensional latent space, and a 0.2 dropout probability— was trained for 500 epochs using an Adam optimizer (learning‑rate $=4\times 10^{-4}$) and a Mean Squared Error (MSE) objective. The weight configuration yielding the lowest validation loss in each fold was retained, producing ten models that were subsequently applied to the independent test set. Imputation performance was then quantified with the metrics reported in Section \ref{sec:metrics}.

\subsection{Upsampling}
After the imputation process, the upsampling is applied to restore the imputed data to its original sequence length, ensuring alignment between the temporal resolution of the imputed signals and that of the original SPEM sequences. A cubic interpolation was used for the upsampling process, specifically following the PCHIP method. The upsampling process increases the sequence length to the original one (i.e., from 500 samples back to 15,000), effectively reconstructing the time series with a higher temporal resolution. However, while upsampling restores the sequence length, fine-grained high-frequency temporal details lost during downsampling cannot be recovered. To address this limitation, a refinement step is incorporated to mitigate the loss of information.

\subsection{Refinement}
To address the loss of fine-grained temporal details incurred during downsampling, a refinement step was incorporated into the processing pipeline using an RAE. 

The RAE enhances the interpolated signals by learning intrinsic temporal patterns, thereby significantly improving the accuracy and fidelity of the reconstructed signals. The RAE achieves this by compressing and reconstructing input sequences through sequential convolutional layers, effectively capturing both local and global temporal dependencies. The model was trained exclusively on sequences without missing values, ensuring it learned accurate representations for minimising reconstruction errors and correcting distortions introduced in prior pipeline stages. Specifically, sequences from 137 participants formed the training set, while sequences from 35 participants comprised the test set. After filtering out sequences containing NaN values, the concatenation of the different SPTs resulted in 2,509 training and 604 test sequences. The training objective was to minimise the reconstruction error, allowing the model to correct distortions introduced in prior pipeline stages.

\begin{figure}[!t]
  \centering
  \includegraphics[scale=0.6]{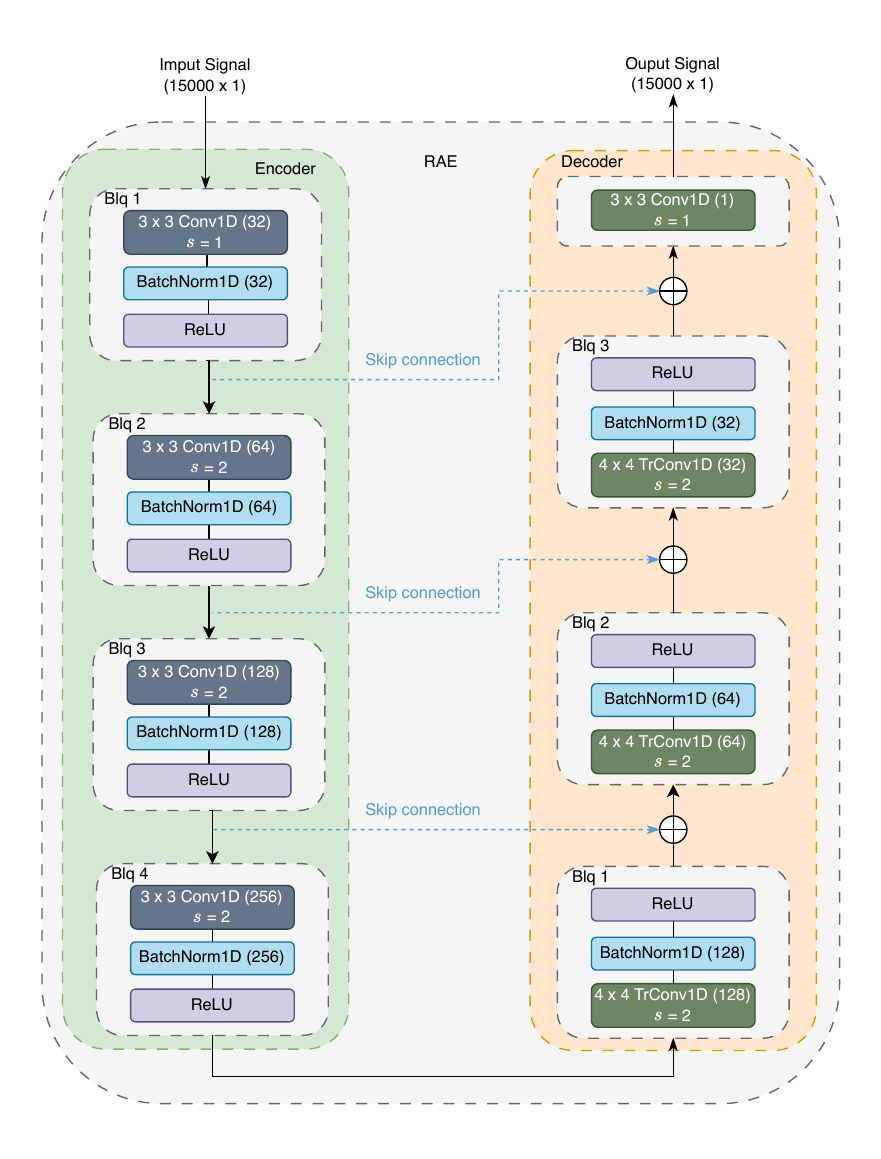} 
  \caption{Details of the architecture of the RAE.}
  \label{fig:fig5}
\end{figure}

The RAE architecture comprises an encoder-decoder structure built from multiple one-dimensional convolutional layers, batch normalisation, and ReLU activation functions. Skip connections between corresponding encoder and decoder layers were implemented to enhance the flow and retention of temporal information throughout the reconstruction process, as they are successfully applied in U-Net-like architectures. Fig. \ref{fig:fig5} details the RAE architecture, explicitly specifying configurations such as the number of convolutional filters, kernel sizes, strides, padding, and activation functions.

Training was carried out using the MSE loss function to quantify the reconstruction error between the original and the reconstructed signal. An Adam optimiser was used for training with a learning rate of $1 \times 10^{-4}$ and a weight decay of $1\times 10^{-5}$. Additionally, a ReduceLROnPlateau learning rate scheduler was used to reduce the learning rate by a factor of 0.5 after 20 consecutive epochs without validation loss improvement. Early stopping with a patience threshold of 50 epochs was also incorporated to mitigate overfitting risks. The training spanned a maximum of 100 epochs, with model parameters corresponding to the lowest validation loss retained for evaluation. A 10-fold cross-validation scheme was followed for validation, partitioning the training dataset (2,509 sequences) into 80\% for training and 20\% for validation for each fold.

\subsection{Evaluation}
This section presents a statistical analysis of the characteristics of the missing data and the metrics used in this paper to objectively evaluate and compare the results. 

\subsubsection{Missing Data Characteristics and Generation of Artificial Blinks}\label{sec:gen_blinks}

A ground truth is required to evaluate and compare the performance of the methods. To achieve this, artificial missing values were inserted into the SPEM sequences in a controlled manner, simulating realistic scenarios. To do so, a statistical characterization of real blinks was carried out, featuring them from an analysis of all sequences in the dataset (encompassing both eyes and all SPTs).

Missing data mechanisms are generally classified into three categories based on the relationship between the missingness and the data itself  \cite{b28},  \cite{b29}: i) Missing Completely at Random (MCAR), where missingness occurs entirely at random and is unrelated to any other variables in the dataset; ii) Missing at Random (MAR), where missingness may be systematically related to other measured variables in the dataset; and iii) Missing Not at Random (MNAR), where missingness is related to the value of the variable itself or other unobserved factors  \cite{b30},  \cite{b31},  \cite{b32},  \cite{b33},  \cite{b26}. Given that blink events in eye movement data are influenced by physiological factors rather than occurring randomly, this study focuses on simulating missing data aligned with the MNAR mechanism, reflecting the characteristics of real blink events. 

The following sections present the method followed to empirically characterise the real blinks and the procedure followed to generate the artificial ones. 

\paragraph{Empirical Characteristics of Real Blinks}
Three parameters were extracted from real blinks to characterise them statistically: i) Duration: the number of missing consecutive samples per blink; ii) Position: the starting position of each blink within the signal; and iii) Number of blinks: the total number of blinks per sequence.

\begin{figure}[!t]
  \centering
  \includegraphics[scale=0.8]{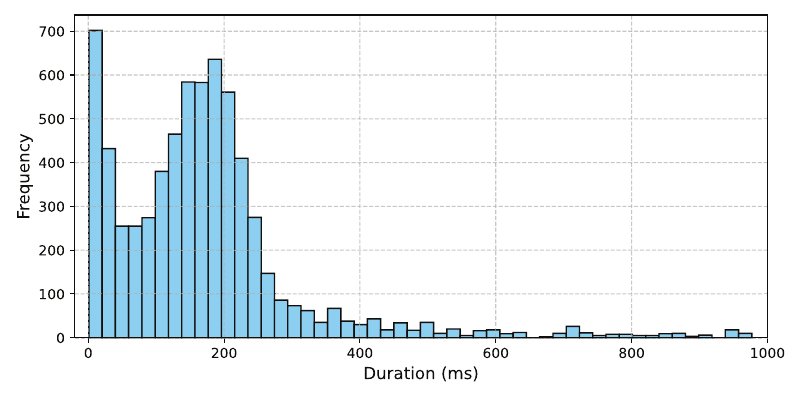} 
  \caption{Histogram corresponding to the duration of the missing values (NaN) for all the sequences in the dataset.}
  \label{fig:fig6}
\end{figure}

The histogram shown in Fig. \ref{fig:fig6}, illustrates the distribution of blink durations. Its non-Gaussian nature, evident in the skewed and irregular shapes of the histogram, indicates that parametric assumptions (e.g., normality) are unsuitable for modelling these characteristics. A similar behaviour is observed for the positions and the total number of blinks, but is not reported for brevity. 

\paragraph{Generation of Artificial Missing Data}
To accurately create artificial missing data reflecting the properties of real blinks, a sampling-based approach was followed. This method ensures that the artificial blinks closely resemble the empirical distributions observed in the dataset of real blinks in terms of duration, position, and frequency. The procedure for generating artificial missing segments involves the following two steps:

\subparagraph{Collection of Empirical Blink Characteristics:}
The empirical distributions of blink durations, positions, and counts were extracted from the dataset. These distributions capture the variability of real blinks with no parametric assumptions.

\subparagraph{Insertion of Artificial Blinks:}
The number of artificial blinks inserted into each signal is randomly selected from the observed blink counts, maintaining the variability and frequency characteristics of the real data. For every artificial blink, both duration and starting position are randomly sampled from the corresponding observed empirical distributions. Boundary checks are then applied: the starting position is adjusted to ensure it is at least 1, and the ending position is capped at the total signal length, with the duration shortened accordingly if needed to avoid exceeding the signal boundary. Artificial blinks are recorded as event annotations, marking the segments designated as missing data (e.g., represented by NaN values) for subsequent analysis.

\subsection{Evaluation Metrics}\label{sec:metrics}

In this section, we summarise the set of quantitative metrics used to evaluate the performance of the imputation methods. The evaluation focuses on both time-domain and frequency-domain characteristics of the sequences to ensure a comprehensive assessment of the imputation quality.

\subsubsection{Time-Domain Metrics}

Time-domain metrics measure the point-wise differences between the original and imputed sequences at the locations where missing data were introduced. The following metrics are computed only at the positions corresponding to the artificially generated missing values.

\paragraph{ Mean Absolute Error (MAE)}
The Mean Absolute Error quantifies the average absolute difference between the original signal $x$ and the imputed signal $\hat{x}$ over the imputed positions, as described in (\ref{eq:mae}):
\begin{equation}
\mathrm{MAE} = \frac{1}{N_{\mathrm{imp}}} \sum_{i\in\mathcal{I}} |x_i - \hat{x}_i| 
\label{eq:mae}
\end{equation}
where $N_{\mathrm{imp}}$ is the total number of imputed samples, and $\mathcal{I}$ denotes the set of indices where imputation was performed.

\paragraph{ Mean Relative Error (MRE)}
The Mean Relative Error assesses the average relative discrepancy between the original and imputed signals:
\begin{equation}
\mathrm{MRE} = \frac{1}{N_{\mathrm{imp}}} \sum_{i\in\mathcal{I}} \left| \frac{x_i - \hat{x}_i}{x_i} \right|
\label{eq:mre}
\end{equation}
To prevent division by zero, samples where $x_i = 0$ are excluded from the computation or handled appropriately.

\paragraph{ Root Mean Square Error (RMSE)}
The Root Mean Square Error measures the standard deviation of the differences between the original and imputed signals:
\begin{equation}
\mathrm{RMSE} = \sqrt{ \frac{1}{N_{\mathrm{imp}}} \sum_{i\in\mathcal{I}} (x_i - \hat{x}_i)^2 }
\label{eq:rmse}
\end{equation}

\paragraph{ Similarity Metric (Sim)}
The Similarity Metric evaluates the linear correlation between the original and imputed signals, analogous to the Pearson correlation coefficient:
\begin{equation}
\mathrm{Sim} = \frac{ \sum_{i\in\mathcal{I}} (x_i - \bar{x})(\hat{x}_i - \bar{\hat{x}}) }
{ \sqrt{ \sum_{i\in\mathcal{I}} (x_i - \bar{x})^2 } \sqrt{ \sum_{i\in\mathcal{I}} (\hat{x}_i - \bar{\hat{x}})^2 } }
\label{eq:sim}
\end{equation}
where $\bar{x}$ and $\bar{\hat{x}}$ are the mean values of the original and imputed signals over the imputed positions, respectively.

\paragraph{ Fraction of Standard Deviation (FSD)}
The Fraction of Standard Deviation normalises the RMSE by the standard deviation of the original signal at the imputed positions, providing a dimensionless measure of error relative to signal variability:
\begin{equation}
\mathrm{FSD} = \frac{\mathrm{RMSE}}{ \sqrt{ \frac{1}{N_{\mathrm{imp}}} \sum_{i\in\mathcal{I}} (x_i - \bar{x})^2 } }
\label{eq:fsd}
\end{equation}

\subsubsection{Frequency-Domain Metrics}

Frequency-domain metrics analyse the impact of imputation on the spectral properties of the signals. These metrics are computed over the entire signal, as frequency analysis requires a complete time series without missing values.

\paragraph{ RMSE in the Frequency Domain (RMSE$_F$)}
$\mathrm{RMSE}_F$ measures the discrepancy between the frequency components of the original and imputed signals:
\begin{equation}
\mathrm{RMSE}_F = \sqrt{ \frac{1}{K} \sum_{k=1}^{K} | X_k - \hat{X}_k |^2 }
\label{eq:rmsef}
\end{equation}
where $X_k$ and $\hat{X}_k$ are the discrete Fourier transform coefficients of the original and imputed signals, respectively, and $K$ is the total number of frequency bins.

\paragraph{ RMSE in Low Frequencies (RMSE$_{F_\mathrm{Low}}$)}
This metric focuses on the preservation of low-frequency components (cutoff: 1~Hz), which are essential for capturing the fundamental features of physiological signals.

\paragraph{ RMSE in High Frequencies (RMSE$_{F_\mathrm{High}}$)}
$\mathrm{RMSE}_{F_\mathrm{High}}$ assesses the accuracy of imputation in the high-frequency range (cutoff: 5~Hz), associated with rapid signal fluctuations and noise.\\

Low values of MAE, MRE, and RMSE indicate high accuracy in reconstructing missing data. High Sim values suggest a strong linear relationship between the original and imputed signals. The FSD provides a normalized error measure relative to signal variability. In the frequency domain, lower $\mathrm{RMSE}_F$, $\mathrm{RMSE}_{F_\mathrm{Low}}$, and $\mathrm{RMSE}_{F_\mathrm{High}}$ values indicate better preservation of the signal's spectral characteristics across frequency bands.

\section{RESULTS}

This section presents the results of the SAITS imputation process and compares the results obtained with those obtained using other state-of-the-art methods, namely: PHCIP, SSA and k-nearest neighbours (KNN). The computed metrics provide quantitative evidence of the performance of the imputation methods in three scenarios: a) imputation of the downsampled sequences; b) upsampled imputed sequences; and, c) upsampled imputed sequences after the RAE (i.e., the complete pipeline). The results for these three scenarios are presented next. For a clearer presentation, a suffix was added to each imputation method to identify the specific scenario (-D, -U and -RAE, respectively).

\subsection{Performance evaluation of the Imputation Block}

The performance of the imputation block was first assessed for the downsampled sequences. Thus, no upsampling and no RAE were applied. Table \ref{tab:downsampled_performance} presents all the time and frequency metrics for each imputation method (i.e., PCHIP-D, SSA-D, KNN-D, and SAITS-D), illustrating the SAITS’s capability to accurately impute missing values in the downsampled sequences. 

\begin{table*}[!t]
\centering
\caption{Performance Metrics for Different Imputation Methods and for Downsampled Sequences}
\label{tab:downsampled_performance}
\begin{tabular}{lcccccccc}
\toprule
\textbf{Method} & \textbf{MAE} & \textbf{MRE} & \textbf{RMSE} & \textbf{Sim} & \textbf{FSD} & $\mathbf{\text{RMSE}_F}$ & $\mathbf{\text{RMSE}_{F_\text{Low}}}$ & $\mathbf{\text{RMSE}_{F_\text{High}}}$ \\
\midrule
PCHIP-D   & 1.92 & 15.48 & 2.89 & 0.79 & 12.66 & 20.76 & 77.76 & 19.31 \\
SSA-D     & 0.14 & 0.54  & 0.18 & 0.83 & 0.93  & 2.32  & 7.91  & 2.18  \\
KNN-D     & 0.12 & 0.71  & 0.14 & 0.84 & 1.90  & 1.99  & 6.14  & 1.91  \\
SAITS-D   & 0.10 & 0.60  & 0.14 & 0.84 & 1.14  & 2.21  & 7.29  & 2.11  \\
\bottomrule
\end{tabular}
\end{table*}

Among the four methods, SAITS-D achieves the lowest point-wise errors ($\mathrm{MAE} = 0.10$, $\mathrm{RMSE} = 0.14$) and the highest similarity to the ground truth ($\mathrm{Sim} = 0.84$), confirming the best performance in the time domain. In contrast, KNN-D records the smallest spectral distortion ($\mathrm{RMSE}_F = 1.99$ vs.\ $2.21$ for SAITS-D) and slightly better preservation of both low- and high-frequency contents. KNN better reproduces the original spectral envelope but leaves marginally larger point-wise discrepancies. Overall, results indicate a trade-off: SAITS optimizes temporal fidelity, whereas KNN offers finer spectral preservation in downsampled SPEM signals.

\subsection{Performance evaluation of the Refinement Autoencoder}	
Across the ten-fold cross‑validation, the RAE provided an average MSE of $1.63 \times 10^{-3}$ for the test set. In contrast, upsampling with a simple cubic interpolation alone produced an MSE of $3.30 \times 10^{-3}$, confirming that the refinement stage halved the reconstruction error. 

\begin{figure}[!t]
  \centering
  \includegraphics[scale=0.5]{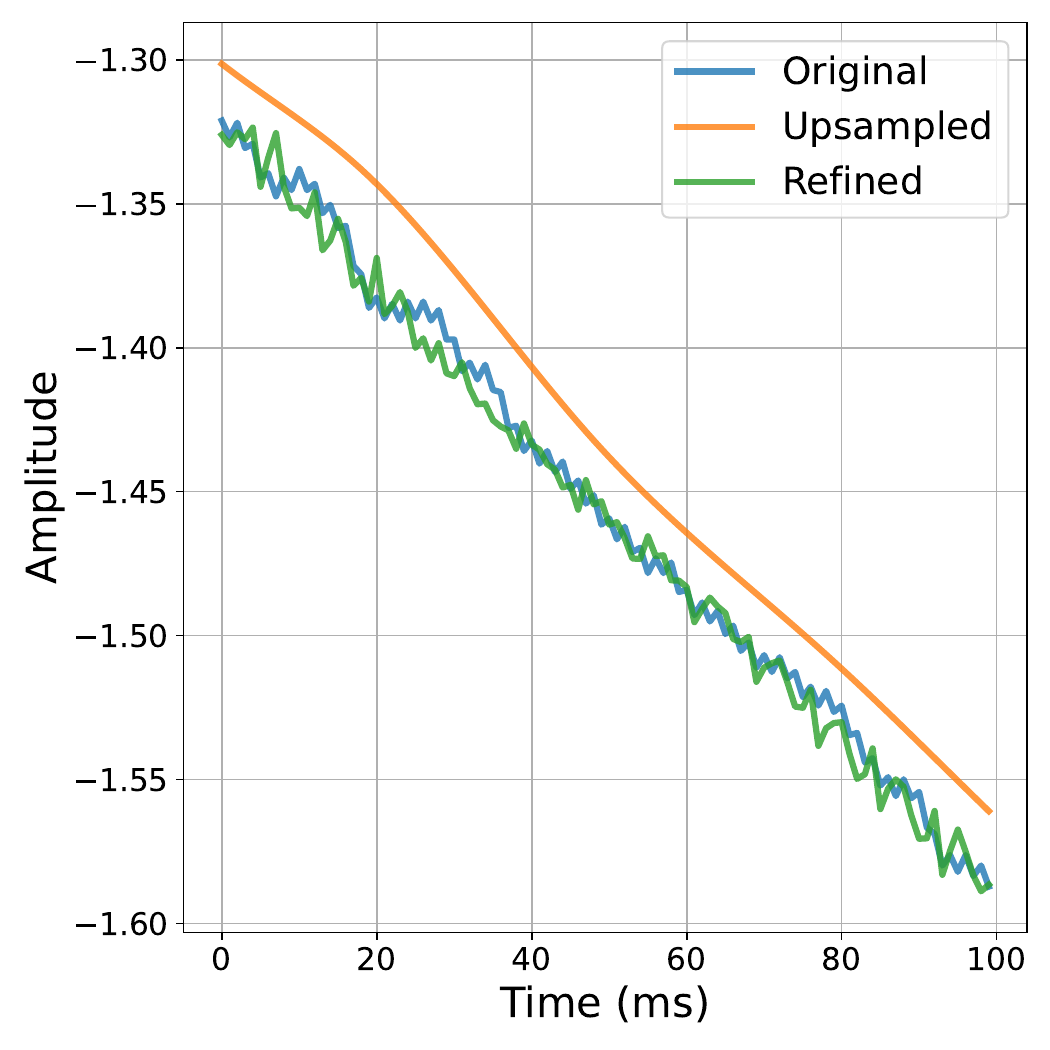} 
  \caption{Comparison of Original, Upsampled, and RAE-Refined Signal Segment (Samples 4100–4200).}
  \label{fig:fig7}
\end{figure}

\begin{figure}[ht]
  \centering
   a{\includegraphics[scale=0.3]{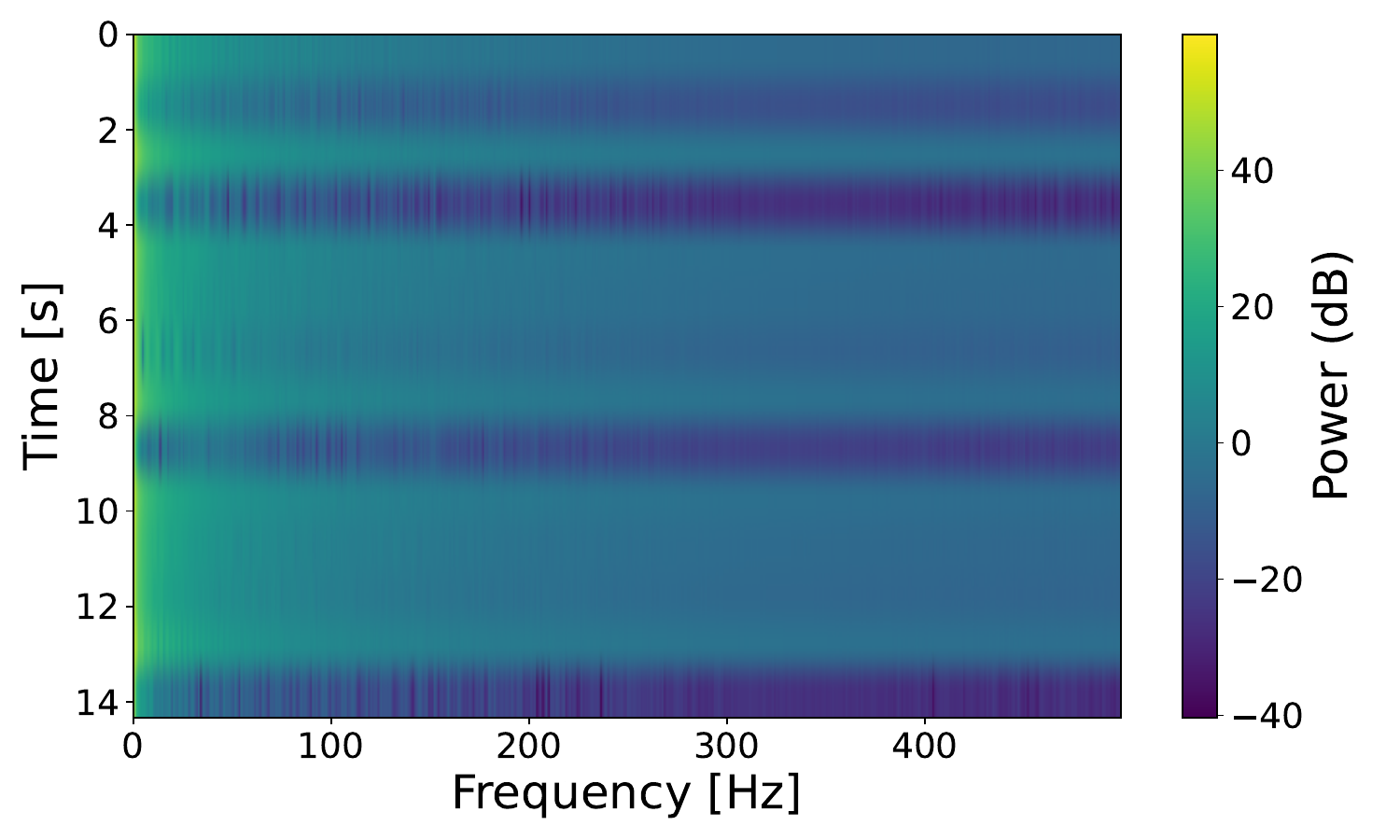}\label{fig:spt_a}}\\[1ex]
   b{\includegraphics[scale=0.3]{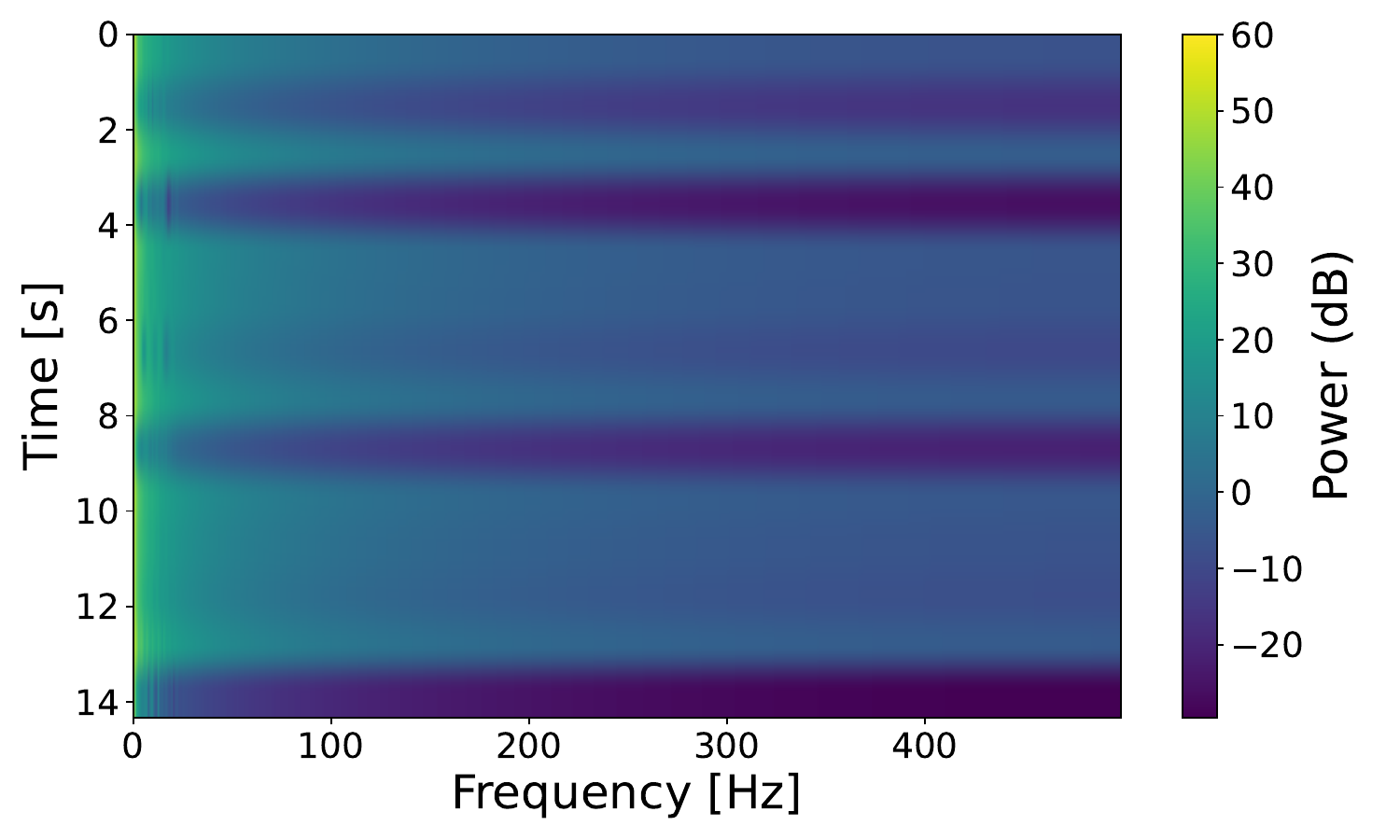}\label{fig:spt_b}}\\[1ex]
   c{\includegraphics[scale=0.3]{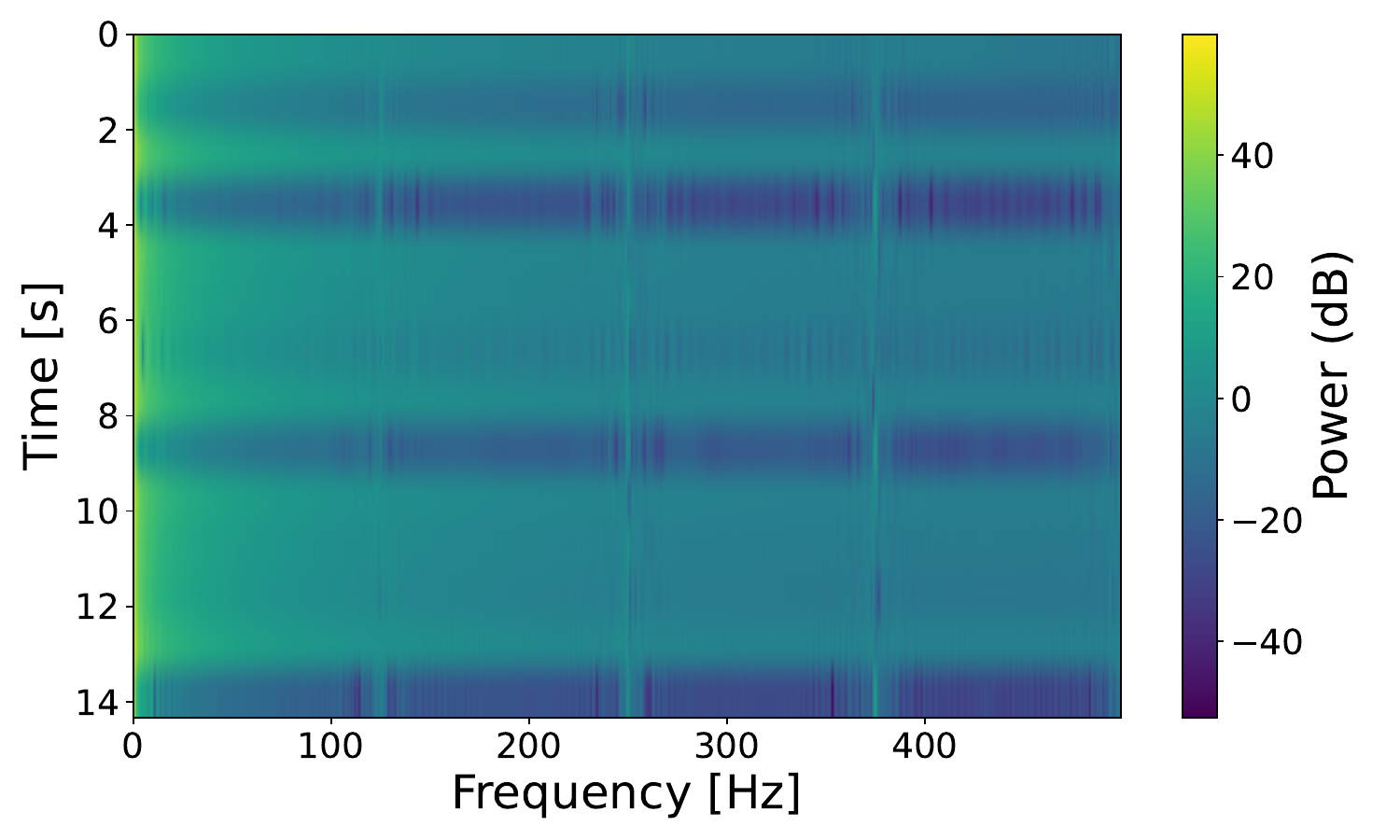}}\label{fig:spt_c}
  \caption{Spectrograms of (a) Original sequence; (b) Upsampled; and (c) RAE-Refined.}
  \label{fig:fig8}
\end{figure}

The effects of the RAE are graphically illustrated in Fig. \ref{fig:fig7}, which presents an illustrative example comparing the original sequence, the upsampled version, and the refined output generated by the RAE. On the other hand, the spectrograms in Fig. \ref{fig:fig8} illustrate the effectiveness of the RAE to recover the spectral characteristics lost during the downsampling process. The original spectrum (Fig. \ref{fig:fig8}a) exhibits well-defined periodic structures and concentrated energy in the low frequencies. In contrast, the spectrogram corresponding to the upsampled signal (Fig. \ref{fig:fig8}b) shows smoothed and less distinct spectral bands. However, the refined signal produced by the RAE method successfully restores the spectral features (Fig. \ref{fig:fig8}c), being more consistent with the original ones. 

These results demonstrate that the RAE restores the original resolution of the SPEM sequences (i.e., 15000 samples) with markedly higher fidelity than the methods developed without it.

\subsection{Performance Evaluation of the Complete Pipeline}	
The proposed complete pipeline (as illustrated in Fig. \ref{fig:fig4}) was evaluated with and without the RAE. Table \ref{tab:pipeline_performance} summarises the results obtained for each method, namely: PCHIP without RAE (PCHIP-U), PCHIP with RAE (PCHIP-RAE), SSA without RAE (SSA-U), SSA with RAE (SSA-RAE), KNN without RAE (KNN-U), KNN with RAE (KNN-RAE), SAITS without RAE (SAITS-U), and SAITS with RAE (SAITS-RAE). 

\begin{table*}[!t]
\centering
\caption{Performance Metrics for Different Imputation Methods}
\label{tab:pipeline_performance}
\begin{tabular}{lcccccccc}
\toprule
\textbf{Method} & \textbf{MAE} & \textbf{MRE} & \textbf{RMSE} & \textbf{Sim} & \textbf{FSD} & $\mathbf{\text{RMSE}_F}$ & $\mathbf{\text{RMSE}_{F_\text{Low}}}$& $\mathbf{\text{RMSE}_{F_\text{High}}}$ \\
\midrule
PCHIP-U   & 1.77 & 13.93 & 2.70 & 0.82 & 11.13 & 108.65 & 1851.08 & 32.28 \\
PCHIP-RAE & 1.66 & 12.75 & 2.53 & 0.81 & 10.55 & 101.46 & 1722.24 & 31.98 \\
SSA-U     & 0.12 & 0.64  & 0.15 & 0.84 & 2.93  & 15.52  & 213.60  & 7.37  \\
SSA-RAE   & 0.12 & 0.66  & 0.16 & 0.82 & 3.50  & 14.25  & 197.68  & 6.50  \\
KNN-U     & 0.11 & 0.85  & 0.14 & 0.82 & 1.77  & 13.83  & 180.47  & 7.60  \\
KNN-RAE   & 0.11 & 0.84  & 0.14 & 0.83 & 1.77  & 12.40  & 162.11  & 6.66  \\
SAITS-U   & 0.10 & 0.76  & 0.14 & 0.84 & 1.14  & 15.04  & 201.49  & 7.29  \\
SAITS-RAE & 0.10 & 0.71  & 0.13 & 0.84 & 1.11  & 13.72  & 184.24  & 6.43  \\
\bottomrule
\end{tabular}
\end{table*}

The results reported in Table~\ref{tab:pipeline_performance} illustrate that SAITS-U retains the best time-domain performance ($\mathrm{MAE} = 0.10$; $\mathrm{RMSE} = 0.14$), whereas KNN-U edges ahead on $\mathrm{RMSE}_F$ (13.83 vs.\ 15.04 for SAITS-U). 
Besides, the gap between SAITS and KNN in the frequency domain narrows with upsampling, but, SAITS-U maintains a consistent advantage in ($\mathrm{Sim} = 0.84$) and provides a 35.6\%  lower FSD than KNN-U, confirming that its attention-based interpolation scales well when temporal resolution is restored.

 The RAE provides modest yet systematic gains for all methods: errors decrease for PCHIP (by 6\% in MAE), remain almost unchanged for SSA, and drop slightly for both KNN and SAITS. Post-refinement, SAITS-RAE re-establishes the global minimum in MAE (0.10) and RMSE (0.13), while KNN-RAE achieves the best overall spectral score ($\mathrm{RMSE}_F = 12.40$). The remaining 10\% gap in $\mathrm{RMSE}_F$ between SAITS-RAE and KNN-RAE is offset by SAITS-RAE’s superior reconstruction of high-frequency fluctuations ($\mathrm{RMSE}_{F_\mathrm{High}} = 6.43$ vs.\ 6.66 for KNN-RAE). These results indicate that the RAE mainly sharpens high-frequency content—most notably benefiting KNN—while the SAITS architecture still provides the most balanced compromise between temporal accuracy and spectral fidelity once the complete pipeline is applied.
 
\subsection{Imputation Performance with Large Missing Intervals}	
To further evaluate the robustness of the proposed SAITS-RAE method in handling large intervals of missing intervals, an additional experiment was conducted, artificially introducing long missing segments into the SPEM sequences. Specifically, we created artificial missing values spanning a duration of 4 s (from 5000 ms to 9000 ms) within the sequences to simulate potential scenarios where tracking is lost for extended periods due to significant movements of the patient, loss of calibration, or technical issues. Table \ref{tab:large_gap} shows the different performance metrics for the imputation methods considered and for large missing intervals.

\begin{table*}[!t]
\centering
\caption{Performance Metrics for Imputation Methods with Large Missing Intervals}
\label{tab:large_gap}
\begin{tabular}{lcccccccc}
\toprule
\textbf{Method} & \textbf{MAE} & \textbf{MRE} & \textbf{RMSE} & \textbf{Sim} & \textbf{FSD} & $\mathbf{\text{RMSE}_F}$ & $\mathbf{\text{RMSE}_{F_\text{Low}}}$ & $\mathbf{\text{RMSE}_{F_\text{High}}}$ \\
\midrule
PCHIP-U   & 1.69 & 10.26 & 2.86 & 0.34 & 3.42 & 205.45 & 4226.65 & 32.23 \\
PCHIP-RAE & 1.64 & 9.87  & 2.75 & 0.34 & 3.29 & 197.34 & 4062.22 & 34.42 \\
SSA-U     & 0.88 & 3.23  & 1.09 & 0.45 & 1.17 & 73.79  & 1662.37 & 7.30  \\
SSA-RAE   & 0.88 & 3.24  & 1.09 & 0.45 & 1.17 & 73.68  & 1662.16 & 6.72  \\
KNN-U     & 1.34 & 6.01  & 1.52 & -0.28 & 1.63 & 160.36 & 3447.46 & 9.30  \\
KNN-RAE   & 1.34 & 5.97  & 1.52 & -0.28 & 1.63 & 159.97 & 3440.10 & 8.95  \\
SAITS-U   & 0.36 & 1.90  & 0.49 & 0.85  & 0.55 & 36.77  & 702.67  & 7.91  \\
SAITS-RAE & 0.36 & 1.89  & 0.49 & 0.85  & 0.55 & 36.21  & 695.72  & 7.29  \\
\bottomrule
\end{tabular}
\end{table*}

In scenarios with large missing intervals, the SAITS-RAE method maintained the best overall performance. Its relative improvement over the PCHIP, SSA, and KNN methods becomes more pronounced as the size of missing data segments increases. In these cases, the PCHIP-U method exhibited significantly higher errors, with an MAE of 1.69 and an MRE of 10.26, indicating poor reconstruction for extensive missing segments. The addition of the RAE slightly improved the performance (PCHIP-RAE), reducing the MAE to 1.64 and the MRE to 9.87, though these errors remain substantially high. KNN-U performed marginally better than PCHIP-U but still yielded inaccurate results (MAE $= 1.34$, MRE $= 6.01$) and displayed a negative similarity with respect to the ground truth (Sim $= -0.28$); RAE refinement left these scores virtually unchanged.

The SSA-U method outperformed both PCHIP-U and KNN-U, achieving an MAE of 0.88 and an MRE of 3.23, but the RAE did not notably enhance SSA's performance. In contrast, the SAITS method demonstrated remarkable robustness, with an MAE of 0.36 and an MRE of 1.90, providing the most accurate reconstruction of large missing intervals. The dedicated refinement network further polished the output (SAITS-RAE: MAE $= 0.36$, MRE $= 1.89$), while also retaining the highest similarity to the reference signal (Sim $= 0.85$).

The frequency-domain metrics further highlight the superiority of SAITS-RAE in handling large missing intervals. The $\mathrm{RMSE}_F$ for SAITS-RAE was 36.21, substantially lower than that of PCHIP (205.45) and SSA (73.68), indicating superior preservation of the signal's spectral characteristics. Additionally, the $\mathrm{RMSE}_{F_\mathrm{Low}}$ was lowest for SAITS-RAE (695.72), compared to 4062.22, 1662.16, and 3440.10 for PCHIP-RAE, SSA-RAE, and KNN-RAE, respectively.

\section{DISCUSSION}
The comprehensive evaluation of the imputation pipeline presented in this study underscores a clear hierarchy among the tested methodologies (Tables 1–3). Performance assessments conducted across multiple scenarios —ranging from moderate data loss conditions to extensive missing intervals— highlight significant differences in performance among traditional interpolation methods (PCHIP, KNN, SSA) and the proposed SAITS approach.

The PCHIP method consistently showed substantial limitations, particularly when faced with extended intervals of missing data. Its inability to accurately model long-term dependencies resulted in high point-wise errors and substantial spectral distortion, as evidenced by elevated MAE values (exceeding 1.6) and extremely high frequency-domain errors. Although the RAE provided modest error reductions, these incremental improvements were insufficient to overcome the inherent shortcomings of purely local interpolation strategies, reaffirming PCHIP's unsuitability for complex physiological time series like SPEM sequences.

The KNN algorithm, while superior to PCHIP in terms of spectral preservation, notably struggled with accurate temporal reconstruction in long missing intervals. Its reliance on exemplar matching from the training data, while beneficial for spectral fidelity, introduced significant artifacts at the point-wise level, culminating in a negative Sim metric for long missing intervals. Post-processing with RAE marginally enhanced KNN performance, yet the method remained substantially less effective than other advanced alternatives when dealing with large gaps.

SSA presented a more balanced performance profile, achieving lower errors than both PCHIP and KNN in standard scenarios. However, its efficacy markedly declined when confronted with extensive gaps in the data, reflecting its dependence on local temporal windows that constrain long-range extrapolation capabilities. The RAE integration did not significantly improve SSA performance, suggesting limited potential for correcting distortions introduced by extensive interpolation in the frequency domain.

The SAITS model consistently outperformed the aforementioned approaches across all evaluation metrics. By effectively capturing complex temporal dependencies without relying on strictly local patterns or exemplar templates, SAITS achieved significantly lower point-wise MSE errors and maintained superior performance even under challenging scenarios with extensive data loss (evidenced by its MAE and Sim). These results confirm the effectiveness of the transformer architecture, especially in scenarios characterised by large intervals of missing data, as frequently encountered in clinical SPEM datasets.

\begin{figure}[ht]
  \centering
  \includegraphics[scale=0.8]{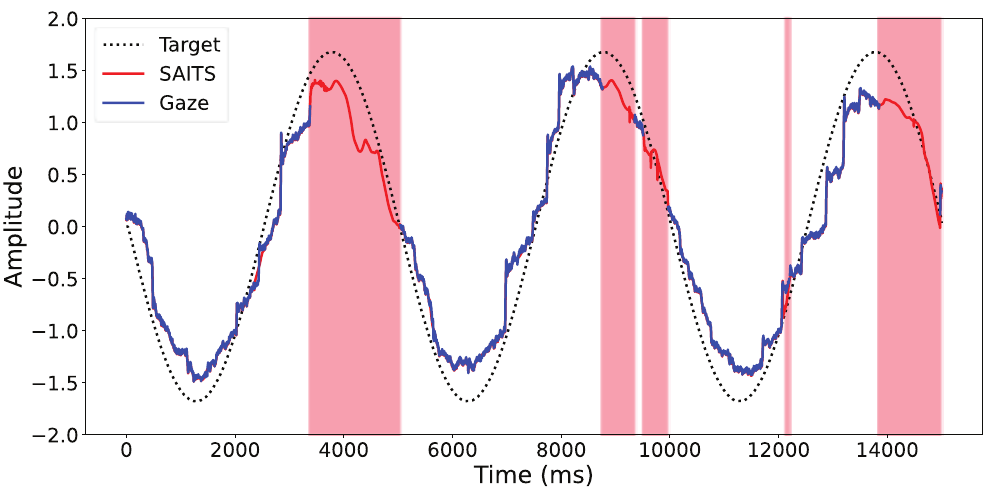} 
  \caption{Illustration of an SPEM sequence (in blue) along with the target (in dotted black) for participant HG032\_V2, SPT12, y-axis and right eye. Red stripes correspond to the missing segments, and imputed data using the method developed is represented in red.}
  \label{fig:fig9}
\end{figure}

\begin{figure}[ht]
  \centering
  \includegraphics[scale=0.8]{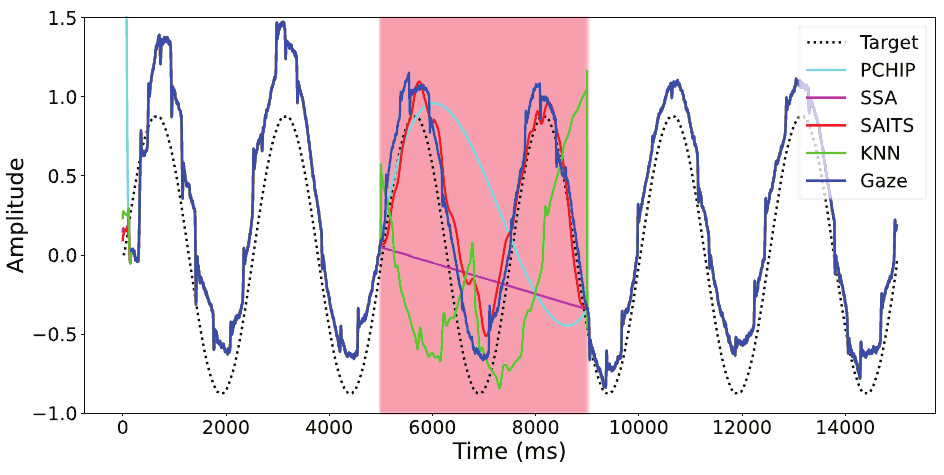} 
  \caption{Reconstruction of SPEM Sequences with Large Missing Intervals Using Different Imputation Methods for participant HG043\_V2, SPT11, x-axis and left eye.}
  \label{fig:fig10}
\end{figure}

Traditional imputation methods tend to perform inadequately when faced with signals exhibiting large contiguous blocks of missing data. In contrast, the proposed SAITS-RAE framework demonstrates robust performance under such challenging conditions. This is exemplified in Fig. \ref{fig:fig9}, which revisits the same SPEM sequence shown in Fig. \ref{fig:fig1}, now with the missing intervals highlighted and the results of the SAITS-RAE interpolation clearly illustrated. For the illustration of all imputation methods in scenarios with large missing intervals, Fig. \ref{fig:fig10} illustrates that the SAITS-RAE method outperforms the other methods, as evidenced by the results presented in Table 3.

The RAE plays a crucial role in enhancing the overall quality of the imputed signals by effectively capturing fine-grained temporal details that are often lost during the upsampling process. By learning intricate temporal dependencies and correcting distortions introduced during interpolation, the RAE ensures that the signals closely approximate the original sequences. This enhancement is evidenced by the consistent improvements across all methods, including the RAE. In particular, for SAITS, the RAE refines both the time-domain and frequency-domain characteristics of the sequences.

Moreover, the RAE demonstrates its effectiveness by consistently providing improvements across all methods. These consistent, albeit sometimes modest gains, suggest that the RAE effectively fine-tunes the initial imputation, capturing subtle signal features that might otherwise be overlooked. Fig. \ref{fig:fig7} and Fig. \ref{fig:fig8} visually show that the RAE significantly enhances the quality of upsampled signals by reconstructing fine-grained temporal patterns and preserving frequency-domain integrity. 

Clinically, these findings carry significant implications. The robust reconstruction performance of SAITS-RAE ensures reliable downstream analysis of SPEM signals, a critical requirement for accurate biomarker extraction. Its resilience in handling both typical and extensive data losses, coupled with its preservation of temporal and spectral integrity, highlights its suitability for clinical and real-world deployment, where data imperfections and intermittent data loss are commonplace.

In summary, the comparative analyses highlight the advantages of the proposed SAITS-RAE approach. This method achieves a significant and consistent improvement in the reconstruction of SPEM sequences. The results underscore SAITS-RAE as a robust, accurate, and clinically viable imputation solution, significantly advancing the quality of SPEM data analysis and interpretation.

\section{CONCLUSIONS}  

In this paper, we introduce a novel framework that combines the SAITS model with an autoencoder-based refinement step to address missing data in SPEM recordings. The proposed method significantly improves the accuracy of reconstructed sequences, outperforming traditional interpolation techniques. This improvement was consistently observed across various scenarios, including those with large missing intervals. 

The SAITS model's ability to handle long-term dependencies and capture intricate temporal relationships makes it particularly well-suited for imputing missing values in SPEM sequences. Its non-autoregressive approach avoids the compounding of errors common in sequential models, and the self-attention mechanism allows it to model complex interactions within the data effectively. By integrating the SAITS model into the proposed pipeline, we leverage its strengths to enhance the robustness and accuracy of missing data reconstruction.

On the other hand, the RAE significantly enhances the quality of the imputation of missing data in SPEM sequences, outperforming traditional methods across various tasks. The combination of SAITS and the RAE provides a robust and effective solution, improving the fidelity of reconstructed signals. This approach holds promise not only for SPEM analysis but also for other biomedical time series data such as EEG and ECG signals, where missing data and long sequences pose significant challenges.

While the method shows significant improvements, future work could explore optimisation techniques to implement more efficient architectures or algorithms. Investigating alternative models or integrating attention mechanisms directly into RAE architectures may further enhance performance. Additionally, exploring strategies for real-time implementation could facilitate immediate analysis in clinical settings, making the approach even more valuable. 

\section*{ACKNOWLEDGMENTS}
This work was supported by the Ministry of Economy and Competitiveness of Spain under Grants PID2021-128469OB-I00 and TED2021-131688B-I00, and by Comunidad de Madrid, Spain. Universidad Politécnica de Madrid supports Julián D. Arias-Londoño through a María Zambrano UP2021-035 grant funded by the European Union-NextGenerationEU.
The authors thank the Neurology Service of Hospital General Universitario Gregorio Marañón, Madrid, for the facilities provided, and Universidad Politécnica de Madrid for providing computing resources on the Magerit Supercomputer. 
Finally, the authors would like to thank the Madrid ELLIS unit (European Laboratory for Learning \& Intelligent Systems) for its indirect support, and to all patients who selflessly participated in the study.

\bibliographystyle{unsrt}
\bibliography{references}

\end{document}